\documentclass{article}
\usepackage{spconf,amsmath,amssymb,graphicx,hyperref}
\usepackage{titlesec}
\usepackage{booktabs}
\usepackage{multirow}
\usepackage{enumitem}
\usepackage[table]{xcolor}
\usepackage{makecell}

\definecolor{lightblue}{rgb}{0.796, 0.894, 0.9808}

\title{CTR-LoRA: Curvature-Aware and Trust-Region Guided Low-Rank Adaptation for Large Language Models}

\name{
\parbox{\textwidth}{\centering
Zhuxuanzi Wang$^{1*}$ \quad Mingqiao Mo$^{1*}$ \quad Xi Xiao$^{2,3}$ \quad Chen Liu$^{4}$ \quad Chenrui Ma$^{5}$ \\
Yunbei Zhang$^{6}$ \quad Xiao Wang$^{3\dagger}$ \quad Smita Krishnaswamy$^{4}$ \quad Tianyang Wang$^{2\dagger}$
}}

\address{$^{1}$ Cornell University \quad
         $^{2}$ University of Alabama at Birmingham \quad
         $^{3}$ Oak Ridge National Laboratory \\
         $^{4}$ Yale University \quad
         $^{5}$ University of California, Irvine \quad
         $^{6}$ Tulane University \\
         {\small $^{*}$Equal contribution \quad $^{\dagger}$Corresponding author: \texttt{tw2@uab.edu, wangx2@ornl.gov}}
}

\begin{document}

\maketitle

\begin{abstract}
Parameter-efficient fine-tuning (PEFT) has become the standard approach for adapting large language models under limited compute and memory budgets. Although previous methods improve efficiency through low-rank updates, quantization, or heuristic budget reallocation, they often decouple the allocation of capacity from the way updates evolve during training. In this work, we introduce CTR-LoRA, a framework guided by curvature trust region that integrates rank scheduling with stability-aware optimization. CTR-LoRA allocates parameters based on marginal utility derived from lightweight second-order proxies and constrains updates using a Fisher/Hessian-metric trust region. Experiments on multiple open-source backbones (7B-13B), evaluated on both in-distribution and out-of-distribution benchmarks, show consistent improvements over strong PEFT baselines. In addition to increased accuracy, CTR-LoRA enhances training stability, reduces memory requirements, and achieves higher throughput, positioning it on the Pareto frontier of performance and efficiency. These results highlight a principled path toward more robust and deployable PEFT.
\end{abstract}


\section{Introduction}
Parameter-efficient fine-tuning (PEFT) has become the standard route to adapt billion-parameter language models under strict compute and memory budgets~\cite{xiao2025visual, li2025magicid, xiao2025visualinstanceawareprompttuning, xiao2025focusfusedobservationchannels, zhang2024dpcore}. Among PEFT designs, low-rank adaptation (LoRA) injects a pair of thin adapters into selected projections and freezes the backbone, achieving strong transfer with \emph{no} inference-time latency and a minimal trainable footprint~\cite{hu2021lora}. Quantization-aware PEFT pushes this boundary further: QLoRA enables 4-bit training and makes even 65B models attainable on a single 48\,GB GPU~\cite{dettmers2023qlora}. Despite its ubiquity, LoRA still faces two pain points: (1) rank schedules are typically uniform or hand-tuned, ignoring the heterogeneous importance of layers and matrices, and (2) training stability depends on update geometry and step size, leaving low-rank updates vulnerable in high-curvature regions. Recent studies also highlight pronounced hyperparameter sensitivity and a trade-off between acquiring new skills and retaining base knowledge~\cite{biderman2024loralearns}.
\begin{figure}[t]
    \centering
    \includegraphics[width=0.98\linewidth]{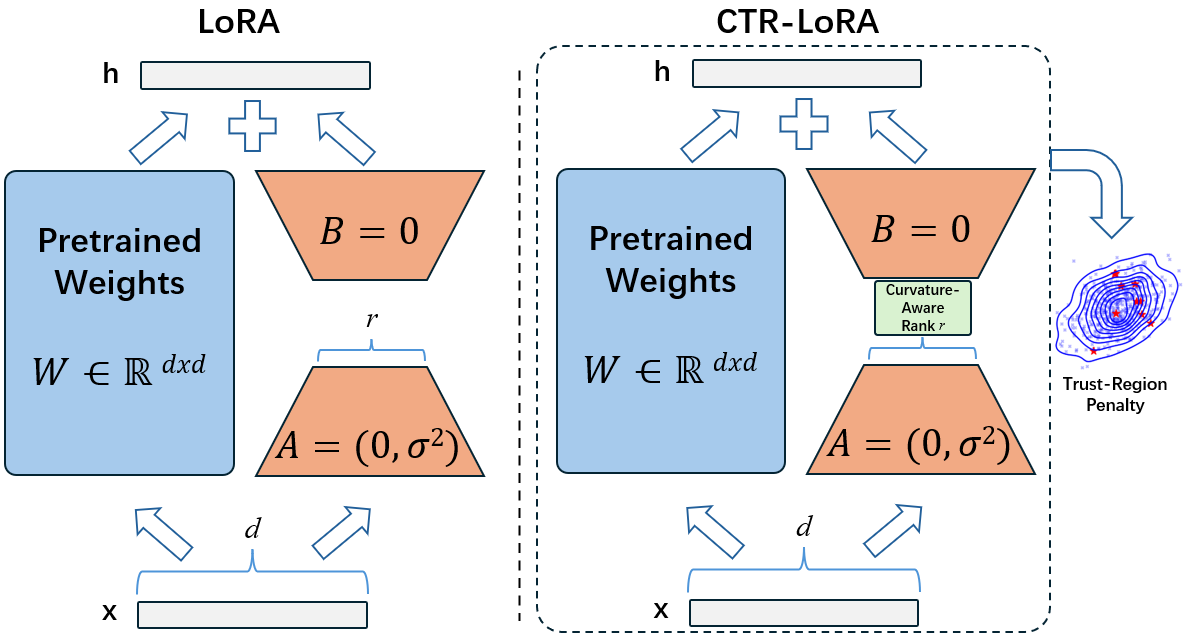}
    \caption{Comparison between LoRA and CTR-LoRA. 
    Left: standard LoRA decomposes the low-rank update $\Delta W=AB^{\top}$ with $A \sim \mathcal{N}(0,\sigma^2)$ and $B=0$, merged into frozen pretrained weights $W \in \mathbb{R}^{d \times d}$. 
    Right: CTR-LoRA introduces curvature-aware rank scheduling (guiding the choice of rank $r$) and a trust-region penalty that regularizes the update trajectory, improving stability and robustness during fine-tuning.}
    \label{fig:lora_ctr_lora}
    \vspace{-0.5em}
\end{figure}
A series of variants tackle these challenges from different angles. \emph{Adaptive-budget} methods redistribute rank across matrices (e.g., AdaLoRA) under a global parameter cap~\cite{zhang2023adalora}. \emph{Optimization-aware} refinements adjust the update rule: DoRA decouples magnitude and direction~\cite{liu2024dora}, while LoRA+ applies asymmetric learning rates to correct imbalanced dynamics~\cite{hayou2024loraplus}. \emph{Initialization-aware} schemes such as PiSSA seed adapters with principal components to accelerate convergence~\cite{meng2024pissa}. More recently, \emph{sensitivity-driven} allocation leverages second-order signals for rank selection with low overhead (Sensitivity-LoRA), underscoring the value of curvature information~\cite{zhang2025sensitivitylora}. Yet most methods still optimize rank \emph{separately} from training dynamics: they decide \emph{where} to place parameters but not \emph{how} low-rank updates evolve across the loss landscape.

\textbf{This work proposes a purely adapter-based solution that unifies \emph{where} to allocate capacity with \emph{how} to move in the parameter space.} We introduce a curvature-informed, trust-region-guided framework that (1) scores marginal utility of rank-1 directions under a global budget for rank scheduling, and (2) regularizes training with a curvature-metric penalty to keep updates stable in high-curvature regions. The approach is plug-and-play with existing LoRA stacks (compatible with LoRA+, DoRA, and 4-bit training), incurs \emph{zero} inference-time cost, and preserves the simplicity practitioners expect. Our contributions are summarized as follows:

\begin{itemize}[leftmargin=14pt,topsep=4pt, itemsep=2pt, parsep=0pt]
\item We introduce a principled criterion based on marginal utility under curvature proxies, replacing uniform or heuristic schedules. This shifts PEFT from manual tuning toward theoretically guided allocation, setting a foundation for more systematic use of limited parameters.
\item We establish a stability-oriented objective that constrains updates in curvature space, ensuring reliable optimization and quantization compatibility. This provides a unifying perspective that links PEFT to trust-region and natural-gradient methods, bridging practice with theory.
\item Our framework is fully compatible with existing LoRA ecosystems, requires no architectural changes, and preserves zero inference overhead. This positions CTR-LoRA not merely as another variant, but as a deployable and extensible paradigm for robust, large-scale adaptation.
\end{itemize}
\vspace{-12pt}

\begin{figure*}[t]
\centering

\includegraphics[width=\linewidth]{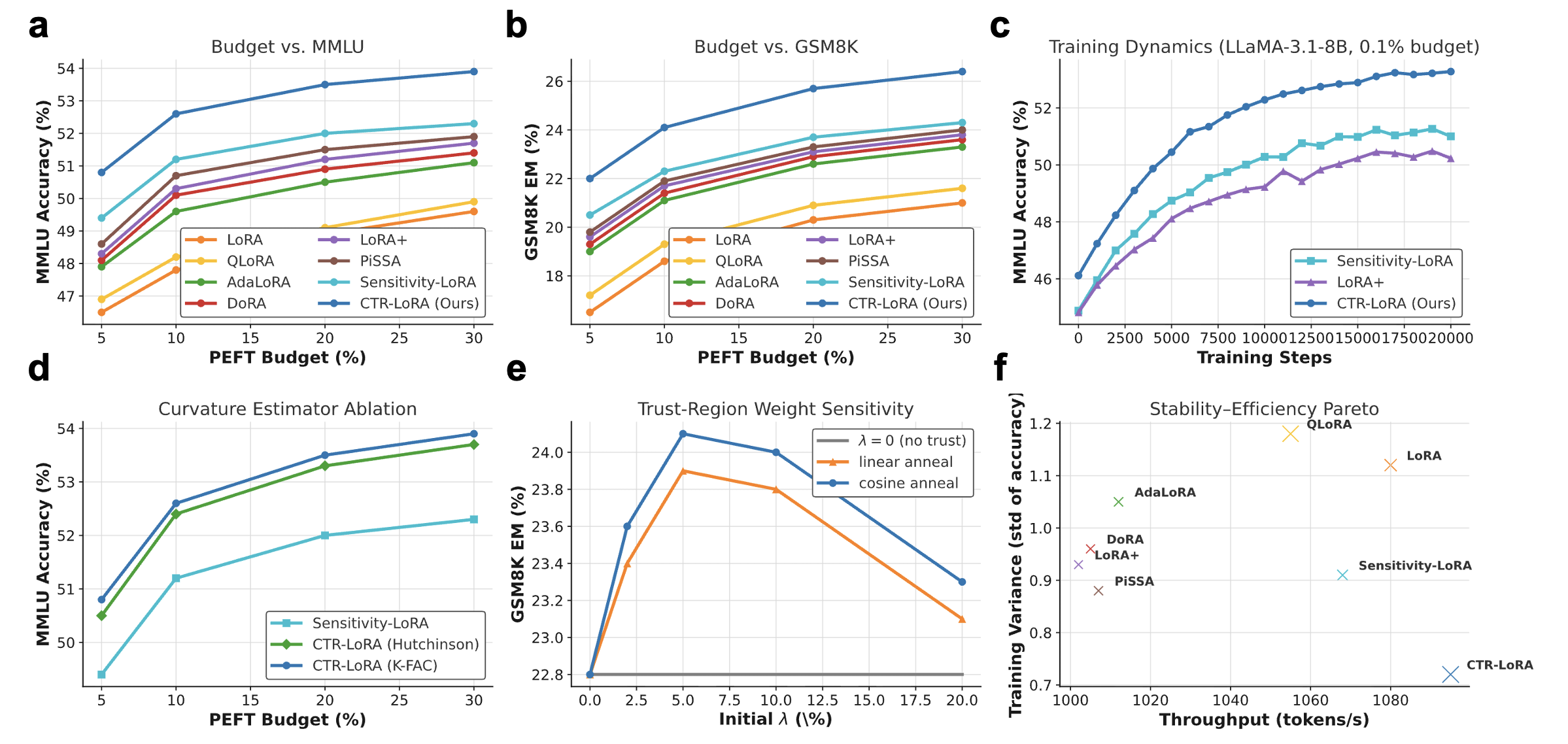}
\vspace{-8pt}
\caption{Comprehensive analyses of CTR-LoRA. 
\textbf{(a-b)} Budget-performance curves show consistent improvements. \textbf{(c)} CTR-LoRA has smoother training dynamics and converges to higher accuracy. \textbf{(d)} K-FAC and Hutchinson proxies outperform sensitivity-only allocation, confirming robustness to curvature estimators. \textbf{(e)} Performance is stable within a moderate trust-region range. \textbf{(f)} CTR-LoRA occupies the Pareto frontier by reducing variance, lowering memory, and increasing throughput.}
\label{fig:sixpanel}
\vspace{-0.5em}
\end{figure*}

\section{Method}

\paragraph{Problem Definition}
We adapt a pretrained transformer with parameter matrices $\{W_\ell\}_{\ell=1}^L$ using low-rank adapters. For each target matrix $W_\ell\in\mathbb{R}^{d_{\text{out}}\times d_{\text{in}}}$, LoRA parameterizes an update $\Delta W_\ell=A_\ell B_\ell^\top$ with rank $r_\ell\ll\min(d_{\text{in}},d_{\text{out}})$ and injects $W_\ell\leftarrow W_\ell+\Delta W_\ell$ while freezing the backbone~\cite{hu2021lora}. We assume a global trainable-parameter budget $B=\sum_{\ell} r_\ell(d_{\text{in}}+d_{\text{out}})$, or equivalently a fixed ratio of base parameters. The central goal of CTR-LoRA is to allocate ranks such that each parameter provides the highest marginal reduction in loss, while simultaneously constraining the update geometry to remain stable in high-curvature regions, all without incurring inference-time latency.

\paragraph{Curvature Proxies Second-Order Signals}
\vspace{-8pt}
Let $\mathcal{L}$ denote the training loss and $G_\ell=\nabla_{W_\ell}\mathcal{L}$ the gradient. To capture local geometry, we approximate curvature with a block-diagonal proxy $M_\ell\succeq 0$ for each matrix. In practice, we employ either Kronecker-factored approximations (K-FAC)~\cite{martens2015kfac}, which decompose the Fisher block as $\mathbb{E}[gg^\top]\otimes \mathbb{E}[aa^\top]$ based on layer inputs $a$ and backpropagated gradients $g$, or Hutchinson-style estimators~\cite{pearlmutter1994fast, avron2011randomized}, which use stochastic Hessian-vector products to approximate diagonals and quadratic forms. These statistics are precomputed on a small calibration set and refreshed intermittently during training. With $M_\ell$ in place, we define the whitened gradient $\tilde{G}_\ell = M_\ell^{-1/2} G_\ell M_\ell^{-1/2}$, which normalizes directions by curvature and aligns with natural-gradient intuition~\cite{amari1998natural}.

\paragraph{Curvature-Aware Rank Scheduling}
\vspace{-8pt}
We evaluate candidate rank-one directions using a local quadratic model. The second-order expansion at $W_\ell$ is given by Eqn~\eqref{eqn:second_order_expansion}.
\begin{align}
\begin{split}
\Delta \mathcal{L} &\approx -\langle G_\ell, \Delta W_\ell\rangle + \tfrac{1}{2}\|\Delta W_\ell\|_{M_\ell}^2\\
\qquad 
\|X\|_{M_\ell}^2 &= \langle X, M_\ell X\rangle
\label{eqn:second_order_expansion}
\end{split}
\end{align}
Confining $\Delta W_\ell$ to a form $\alpha UV^\top$ with $\|U\|_2=\|V\|_2 = 1$, the optimal step size is $\alpha^\star=\langle G_\ell, UV^\top\rangle/\|UV^\top\|_{M_\ell}^2$, yielding a predicted decrease as shown in Eqn~\eqref{eqn:predicted_decrease}.
\begin{align}
\begin{split}
\mathcal{D}_\ell(U,V) &= \tfrac{1}{2}\frac{\langle G_\ell, UV^\top\rangle^2}
{\|UV^\top\|_{M_\ell}^2} \\
&= \tfrac{1}{2}\,\sigma^2(M_\ell^{-1/2} G_\ell M_\ell^{-1/2})
\label{eqn:predicted_decrease}
\end{split}
\end{align}
Here, the equality holds when $(U,V)$ corresponds to a singular pair of the whitened gradient. Thus the leading singular vectors of $\tilde{G}_\ell$ naturally provide descending marginal utilities. To respect the global budget $B$, we collect candidate directions across layers and select them greedily by their marginal utility scores. Each selection assigns one unit of rank, optionally followed by rank-one deflation to update the residual spectrum. In practice, randomized SVD~\cite{halko2011finding} with small oversampling suffices to approximate these spectra efficiently. Compared with uniform or hand-tuned schedules, this strategy explicitly favors matrices where curvature-adjusted gradients carry the most useful information.

\paragraph{Trust-Region PEFT Objective}
\vspace{-8pt}
Rank allocation alone does not constrain how updates traverse the loss landscape. To further stabilize training, we augment the objective with a curvature-metric regularizer (Eqn~\eqref{eqn:regularizer}).
\begin{equation}
\min_{\{A_\ell,B_\ell\}} \; \mathcal{L}_{\text{task}} + \lambda\sum_{\ell=1}^L \|A_\ell B_\ell^\top\|_{M_\ell}^2
\label{eqn:regularizer}
\end{equation}
This term enforces a trust region in the ellipsoidal norm induced by $M_\ell$, damping unstable directions and improving quantization-friendliness. The additional cost per batch is linear in adapter size, as it only requires quadratic forms. We anneal $\lambda$ toward zero during training, ensuring strong regularization early for stability and relaxed constraints later for expressiveness.


\paragraph{Complexity and Memory}
\vspace{-8pt}
For each matrix, K-FAC maintains two covariance factors whose inverses are computed efficiently by eigendecomposition or Cholesky. Randomized SVD with target rank $k_\ell$ adds cost $\mathcal{O}(d_{\text{out}}d_{\text{in}}k_\ell)$ with small constants, while curvature refreshes every few hundred steps amortize overhead. At inference, adapters are merged into the base model, preserving zero latency. When $M_\ell$ is the Fisher block, our scoring coincides with selecting dominant natural-gradient modes, and the trust-region regularizer reduces to Tikhonov regularization in this metric~\cite{amari1998natural}. Compared with AdaLoRA~\cite{zhang2023adalora} and Sensitivity-LoRA~\cite{zhang2025sensitivitylora}, CTR-LoRA explicitly couples rank placement with update geometry, achieving higher stability and memory efficiency under identical budgets.

\begin{table*}[t]
\centering
\scriptsize
\setlength{\tabcolsep}{4pt}
\renewcommand{\arraystretch}{1}
\caption{Comparison of parameter-efficient fine-tuning methods across four open LLM backbones (LLaMA-3.1-8B, Mistral-7B, LLaMA-2-13B, Qwen2.5-7B). Results are reported as accuracy (\%) on four representative benchmarks.}
\label{tab:main_all_compact}
\resizebox{\textwidth}{!}{%
\begin{tabular}{l cccc cccc cccc cccc}
\toprule
\multirow{2}{*}{Method} &
\multicolumn{4}{c}{\textbf{LLaMA-3.1-8B}} &
\multicolumn{4}{c}{\textbf{Mistral-7B}} &
\multicolumn{4}{c}{\textbf{LLaMA-2-13B}} &
\multicolumn{4}{c}{\textbf{Qwen2.5-7B}} \\
\cmidrule(lr){2-5}\cmidrule(lr){6-9}\cmidrule(lr){10-13}\cmidrule(lr){14-17}
& \texttt{MMLU} & \texttt{GSM8K} & \texttt{ARC-C} & \texttt{HellaSwag} & \texttt{MMLU} & \texttt{GSM8K} & \texttt{ARC-C} & \texttt{HellaSwag} & \texttt{MMLU} & \texttt{GSM8K} & \texttt{ARC-C} & \texttt{HellaSwag} & \texttt{MMLU} & \texttt{GSM8K} & \texttt{ARC-C} & \texttt{HellaSwag} \\
\midrule
LoRA~\cite{hu2021lora}                   & 47.8 & 18.6 & 45.2 & 77.5 & 46.9 & 17.9 & 44.8 & 76.8 & 49.1 & 19.4 & 46.0 & 78.0 & 48.3 & 18.7 & 45.5 & 77.2 \\
QLoRA~\cite{dettmers2023qlora}           & 48.2 & 19.3 & 45.9 & 78.0 & 47.5 & 18.8 & 45.4 & 77.4 & 49.6 & 20.0 & 46.5 & 78.5 & 48.9 & 19.4 & 46.1 & 77.8 \\
AdaLoRA~\cite{zhang2023adalora}          & 49.6 & 21.1 & 46.5 & 78.8 & 48.7 & 20.2 & 45.9 & 77.9 & 50.4 & 21.2 & 47.2 & 79.1 & 50.1 & 20.7 & 46.8 & 78.6 \\
DoRA~\cite{liu2024dora}                  & 50.1 & 21.4 & 46.9 & 79.1 & 49.0 & 20.8 & 46.2 & 78.2 & 50.8 & 21.6 & 47.6 & 79.4 & 50.5 & 21.0 & 47.1 & 78.9 \\
LoRA+~\cite{hayou2024loraplus}           & 50.3 & 21.7 & 47.2 & 79.5 & 49.4 & 21.0 & 46.5 & 78.6 & 51.2 & 22.0 & 47.8 & 79.7 & 50.9 & 21.3 & 47.4 & 79.2 \\
PiSSA~\cite{meng2024pissa}               & 50.7 & 21.9 & 47.3 & 79.6 & 49.8 & 21.2 & 46.7 & 78.8 & 51.5 & 22.2 & 48.1 & 79.9 & 51.3 & 21.6 & 47.6 & 79.4 \\
Sensitivity-LoRA~\cite{zhang2025sensitivitylora} 
                                         & 51.2 & 22.3 & 47.9 & 79.9 &  50.3 &  21.8 &  47.2 &  79.1 &  52.0 &  22.6 &  48.4 & 80.0 &  51.8 &  22.1 &  48.0 & 79.7 \\
\rowcolor{lightblue!30}
\textbf{CTR-LoRA (ours)}                 & \textbf{52.6} & \textbf{24.1} & \textbf{49.2} & \textbf{80.8} & \textbf{51.8} & \textbf{23.5} & \textbf{48.7} & \textbf{80.1} &  \textbf{53.3} & \textbf{24.3} & \textbf{49.8} & \textbf{81.0} & \textbf{53.0} & \textbf{23.8} & \textbf{49.1} & \textbf{80.5} \\
\bottomrule
\end{tabular}
}
\vspace{-12pt}
\end{table*}

\begin{table}[t]
\centering
\scriptsize
\setlength{\tabcolsep}{4pt}
\renewcommand{\arraystretch}{1}
\caption{Accuracies (\%) on out-of-distribution benchmarks (\texttt{BoolQ} / \texttt{WinoGrande}) under a 0.1\% PEFT budget.}
\label{tab:ood}
\resizebox{0.46\textwidth}{!}{%
\begin{tabular}{l cc cc cc cc}
\toprule
\multirow{2}{*}{Method} &
\multicolumn{2}{c}{\makecell{\textbf{LLaMA}\\3.1-8B}} &
\multicolumn{2}{c}{\makecell{\textbf{Mistral}\\7B}} &
\multicolumn{2}{c}{\makecell{\textbf{LLaMA}\\2-13B}} &
\multicolumn{2}{c}{\makecell{\textbf{Qwen}\\2.5-7B}} \\
\cmidrule(lr){2-3}\cmidrule(lr){4-5}\cmidrule(lr){6-7}\cmidrule(lr){8-9}
& \texttt{BoolQ} & \texttt{WG} & \texttt{BoolQ} & \texttt{WG} & \texttt{BoolQ} & \texttt{WG} & \texttt{BoolQ} & \texttt{WG} \\
\midrule
LoRA~\cite{hu2021lora}          & 70.2 & 65.8 & 69.7 & 65.2 & 71.5 & 66.9 & 70.8 & 66.1 \\
QLoRA~\cite{dettmers2023qlora}  & 70.9 & 66.3 & 70.4 & 65.9 & 72.1 & 67.3 & 71.5 & 66.6 \\
AdaLoRA~\cite{zhang2023adalora} & 72.3 & 67.1 & 71.5 & 66.8 & 73.0 & 68.0 & 72.5 & 67.2 \\
DoRA~\cite{liu2024dora}         & 72.8 & 67.6 & 71.9 & 67.2 & 73.4 & 68.5 & 72.9 & 67.7 \\
LoRA+~\cite{hayou2024loraplus}  & 73.0 & 67.9 & 72.2 & 67.5 & 73.8 & 68.7 & 73.2 & 68.0 \\
PiSSA~\cite{meng2024pissa}      & 73.2 & 68.0 & 72.5 & 67.8 & 74.1 & 68.9 & 73.5 & 68.2 \\
Sensitivity-LoRA~\cite{zhang2025sensitivitylora}
                                & 73.8 & 68.4 & 73.0 & 68.2 & 74.6 & 69.2 & 74.0 & 68.6 \\
\rowcolor{lightblue!30}
\textbf{CTR-LoRA (ours)}        & \textbf{75.1} & \textbf{69.7} & \textbf{74.2} & \textbf{69.4} & \textbf{76.0} & \textbf{70.3} & \textbf{75.5} & \textbf{69.8} \\
\bottomrule
\end{tabular}
}
\end{table}

\begin{table}[t]
\centering
\scriptsize
\setlength{\tabcolsep}{4pt}
\renewcommand{\arraystretch}{1}
\caption{Ablations on CTR-LoRA components with LLaMA-3.1-8B at 0.1\% budget. Accuracies (\%) are reported.}
\label{tab:ablation_components}
\resizebox{0.46\textwidth}{!}{%
\begin{tabular}{lccccc}
\toprule
Variant & \texttt{MMLU} & \texttt{GSM8K} & \texttt{ARC-C} & \texttt{HellaSwag} & \textbf{Average} \\
\midrule
CTR-LoRA (full)                        & 52.6 & 24.1 & 49.2 & 80.8 & 51.7 \\
\quad w/o trust region ($\lambda{=}0$) & 51.3 & 22.9 & 48.3 & 80.0 & 50.6 \\
\quad uniform rank (no sched.)         & 51.5 & 23.0 & 48.0 & 79.9 & 50.6 \\
\quad w/o both (no trust \& uniform)   & 50.6 & 22.1 & 47.5 & 79.4 & 49.9 \\
\midrule
\quad + LoRA$+$ (asym.\ LR)            & 52.9 & 24.3 & 49.4 & 80.9 & 51.9 \\
\quad + DoRA (mag/dir decouple)        & 53.0 & 24.4 & 49.5 & 81.0 & 52.0 \\
\quad + PiSSA (SVD init)               & \textbf{53.1} & \textbf{24.6} & \textbf{49.6} & \textbf{81.1} & \textbf{52.1} \\
\bottomrule
\end{tabular}
}
\vspace{-12pt}
\end{table}

\vspace{-12pt}
\section{Experiments}

\paragraph{Experimental Setup}
\vspace{-8pt}
We evaluate CTR-LoRA on several open-source LLMs to ensure reproducibility and scalability. Our main backbones are LLaMA-3.1-8B and Mistral-7B, with additional results on LLaMA-2-13B and Mixtral-8x7B to study scaling. All models are taken from HuggingFace Transformers, with frozen backbones and only adapter parameters updated.  Experiments cover a broad set of benchmarks: \texttt{MMLU}~\cite{hendrycks2021mmlu} (5-shot), \texttt{GSM8K}~\cite{cobbe2021gsm8k} (8-shot), \texttt{ARC-C}~\cite{clark2018arc} (25-shot), and \texttt{HellaSwag}~\cite{zellers2019hellaswag} (10-shot), as well as \texttt{BoolQ}~\cite{clark2019boolq} (zero-shot) and \texttt{WinoGrande}~\cite{sakaguchi2021WinoGrande} for out-of-distribution evaluation. We compare against representative PEFT baselines, including LoRA~\cite{hu2021lora}, QLoRA~\cite{dettmers2023qlora}, AdaLoRA~\cite{zhang2023adalora}, DoRA~\cite{liu2024dora}, LoRA+~\cite{hayou2024loraplus}, PiSSA~\cite{meng2024pissa}, and Sensitivity-LoRA~\cite{zhang2025sensitivitylora}. All methods are trained under matched budgets of 0.1\% and 0.3\% of backbone parameters. Training uses AdamW with learning rate $2\times 10^{-4}$, cosine decay, and 0.03 warm-up. Batch size is 128 with sequence length 2,048, gradient checkpointing, and FlashAttention. For CTR-LoRA, curvature proxies are estimated on a 2k-sample calibration set, refreshed every 500 steps, and the trust-region weight $\lambda$ is annealed from 0.1 to 0. Evaluation follows \texttt{lm-eval-harness}~\cite{gao2021lmeval}, averaging results over three seeds. Besides accuracy, we also report trainable parameter ratio, peak memory, and throughput. Ablations study different curvature estimators (K-FAC vs. Hutchinson), values of $\lambda$, the effect of LoRA+ and DoRA refinements, and budget--performance trade-offs, along with visualizations of rank allocation, curvature spectra, and training stability.

\paragraph{Main Results}
\label{sec:exp}
\vspace{-8pt}
Table~\ref{tab:main_all_compact} reports results on four benchmarks across four open LLM backbones. 
LoRA and QLoRA offer only modest improvements, as they lack mechanisms to adaptively allocate capacity. 
AdaLoRA and DoRA provide incremental gains by redistributing rank or decoupling update magnitude and direction, while LoRA+ and PiSSA accelerate convergence through optimization refinements. 
Sensitivity-LoRA improves further by leveraging second-order information, but its benefits are uneven: \texttt{MMLU} and \texttt{HellaSwag} improve, whereas \texttt{GSM8K} and \texttt{ARC-C} see limited gains, revealing the limitation of sensitivity-only allocation under high-curvature updates. CTR-LoRA consistently outperforms all baselines. 
On LLaMA-2-13B, it surpasses Sensitivity-LoRA by +1.3 on \texttt{MMLU}, +1.7 on \texttt{GSM8K}, +1.4 on \texttt{ARC-C}, and +0.8 on \texttt{HellaSwag}; on Qwen2.5-7B, the margins remain +1.2-1.7. 
These results confirm that coupling curvature-aware rank scheduling with a trust-region constraint is essential: the former determines \emph{where} to allocate parameters, while the latter governs \emph{how} updates evolve, producing both stronger accuracy and greater stability. 
Notably, CTR-LoRA's superiority holds across backbones with different scales and training corpora, highlighting its robustness and generalizability as a practical PEFT strategy.

\paragraph{Out-of-Distribution Generalization}
\vspace{-8pt}
Table~\ref{tab:ood} reports results on \texttt{BoolQ} and \texttt{WinoGrande}, which probe out-of-distribution robustness. Sensitivity-LoRA remains the strongest baseline, confirming the utility of sensitivity-guided allocation, but its margins shrink particularly on \texttt{WinoGrande}, where it is only slightly better than PiSSA. In contrast, CTR-LoRA consistently achieves the highest scores across all models. On \texttt{BoolQ}, it improves over Sensitivity-LoRA by 1.2 to 1.5 points; on \texttt{WinoGrande}, by 1.2 to 1.5 as well. These results highlight that robust generalization in PEFT requires more than rank allocation alone: coupling curvature-aware scheduling with a trust-region constraint stabilizes updates under distributional shift, making CTR-LoRA more reliable for deployment in dynamic real-world settings.

\paragraph{Ablation and Diagnostic Analysis}
\vspace{-8pt}
Table~\ref{tab:ablation_components} evaluates the contribution of each component on LLaMA-3.1-8B under a 0.1\% budget. Removing either curvature-aware scheduling or the trust-region penalty reduces performance by more than one point, and discarding both leads to further degradation, confirming that allocation (\emph{where}) and regulation (\emph{how}) must be coupled. Adding LoRA$+$, DoRA, or PiSSA yields small but consistent gains, with PiSSA giving the best average, showing that CTR-LoRA is compatible with recent refinements. Fig.~\ref{fig:sixpanel} further probes CTR-LoRA from multiple perspectives. Budget-performance curves on \texttt{MMLU} and \texttt{GSM8K} show consistent improvements, particularly under tight budgets and in multi-step reasoning tasks. Training dynamics reveal faster and more stable convergence. Ablations with different curvature proxies (K-FAC vs. Hutchinson) show both outperform sensitivity-only allocation, with minor differences, highlighting robustness. Varying $\lambda$ confirms that a moderate range yields the best results, while overly large values slightly suppress learning. Finally, efficiency analysis places CTR-LoRA on the Pareto frontier, combining lower memory, higher throughput, and reduced variance. These results demonstrate that CTR-LoRA's advantage comes from the synergy of curvature-aware rank scheduling and trust-region regularization, which jointly ensure effective allocation, stable optimization, and broad robustness.

\vspace{-10pt}
\section{Conclusion}
\vspace{-8pt}
We introduced CTR-LoRA, a curvature-trust region guided framework for parameter-efficient fine-tuning of large language models. By combining curvature-aware rank scheduling with a stability-oriented trust-region objective, our method addresses both allocation and optimization challenges inherent in prior PEFT approaches. Experiments on four open backbones and multiple benchmarks, including out-of-distribution settings, show that CTR-LoRA consistently surpasses strong baselines while improving efficiency and stability. These results highlight CTR-LoRA as a principled and practical step toward robust and deployable PEFT, with future opportunities in multi-task and continual adaptation.

\vspace{-12pt}
\section*{Acknowledgments}
\vspace{-8pt}
This manuscript has been co-authored by ORNL, operated by UT-Battelle, LLC under Contract No. DE-AC05-00OR22725 with the U.S. Department of Energy. Any subjective views or opinions that might be expressed in the paper do not necessarily represent the views of the U.S. Department of Energy or the United States Government.

\bibliographystyle{IEEEbib}
{\small
\bibliography{refs}

\begin{thebibliography}{10}

\bibitem{xiao2025visual}
Xi~Xiao, Yunbei Zhang, Yanshuh Li, Xingjian Li, Tianyang Wang, Jihun Hamm, Xiao Wang, and Min Xu,
\newblock ``Visual variational autoencoder prompt tuning,''
\newblock {\em arXiv preprint arXiv:2503.17650}, 2025.

\bibitem{li2025magicid}
Hengjia Li, Lifan Jiang, Xi~Xiao, Tianyang Wang, Hongwei Yi, Boxi Wu, and Deng Cai,
\newblock ``Magicid: Hybrid preference optimization for id-consistent and dynamic-preserved video customization,''
\newblock {\em arXiv preprint arXiv:2503.12689}, 2025.

\bibitem{xiao2025visualinstanceawareprompttuning}
Xi~Xiao, Yunbei Zhang, Xingjian Li, Tianyang Wang, Xiao Wang, Yuxiang Wei, Jihun Hamm, and Min Xu,
\newblock ``Visual instance-aware prompt tuning,'' 2025.

\bibitem{xiao2025focusfusedobservationchannels}
Xi~Xiao, Aristeidis Tsaris, Anika Tabassum, John Lagergren, Larry~M. York, Tianyang Wang, and Xiao Wang,
\newblock ``Focus: Fused observation of channels for unveiling spectra,'' 2025.

\bibitem{zhang2024dpcore}
Yunbei Zhang, Akshay Mehra, Shuaicheng Niu, and Jihun Hamm,
\newblock ``Dpcore: Dynamic prompt coreset for continual test-time adaptation,''
\newblock {\em arXiv preprint arXiv:2406.10737}, 2024.

\bibitem{hu2021lora}
Edward~J. Hu, Yelong Shen, Phillip Wallis, Zeyuan Allen-Zhu, Yuanzhi Li, Shean Wang, Lu~Wang, and Weizhu Chen,
\newblock ``Lora: Low-rank adaptation of large language models,''
\newblock {\em arXiv preprint arXiv:2106.09685}, 2021.

\bibitem{dettmers2023qlora}
Tim Dettmers, Artidoro Pagnoni, Ari Holtzman, and Luke Zettlemoyer,
\newblock ``Qlora: Efficient finetuning of quantized llms,''
\newblock {\em arXiv preprint arXiv:2305.14314}, 2023.

\bibitem{biderman2024loralearns}
Dan Biderman, Jacob Portes, Jose Javier~Gonzalez Ortiz, Mansheej Paul, Philip Greengard, Connor Jennings, Daniel King, Sam Havens, Vitaliy Chiley, Jonathan Frankle, Cody Blakeney, and John~P. Cunningham,
\newblock ``Lora learns less and forgets less,''
\newblock {\em Transactions on Machine Learning Research}, 2024,
\newblock arXiv:2405.09673.

\bibitem{zhang2023adalora}
Qingru Zhang, Minshuo Chen, Alexander Bukharin, Nikos Karampatziakis, Pengcheng He, Yu~Cheng, Weizhu Chen, and Tuo Zhao,
\newblock ``Adalora: Adaptive budget allocation for parameter-efficient fine-tuning,''
\newblock {\em arXiv preprint arXiv:2303.10512}, 2023.

\bibitem{liu2024dora}
Shih-Yang Liu, Chien-Yi Wang, Hongxu Yin, Pavlo Molchanov, Yu-Chiang~Frank Wang, Kwang-Ting Cheng, and Min-Hung Chen,
\newblock ``Dora: Weight-decomposed low-rank adaptation,''
\newblock in {\em Proceedings of the 41st International Conference on Machine Learning (ICML)}. 2024, PMLR.

\bibitem{hayou2024loraplus}
Soufiane Hayou, Nikhil Ghosh, and Bin Yu,
\newblock ``Lora+: Efficient low rank adaptation of large models,''
\newblock {\em arXiv preprint arXiv:2402.12354}, 2024.

\bibitem{meng2024pissa}
Fanxu Meng, Zhaohui Wang, and Muhan Zhang,
\newblock ``Pissa: Principal singular values and singular vectors adaptation of large language models,''
\newblock {\em arXiv preprint arXiv:2404.02948}, 2024.

\bibitem{zhang2025sensitivitylora}
Qingru Zhang, Minshuo Chen, Pengcheng He, Yu~Cheng, Weizhu Chen, and Tuo Zhao,
\newblock ``Sensitivity-lora: Sensitivity-guided rank allocation for low-rank adaptation,''
\newblock {\em arXiv preprint arXiv:2509.09119}, 2025.

\bibitem{martens2015kfac}
James Martens and Roger Grosse,
\newblock ``Optimizing neural networks with kronecker-factored approximate curvature,''
\newblock in {\em ICML}, 2015.

\bibitem{pearlmutter1994fast}
Barak~A. Pearlmutter,
\newblock ``Fast exact multiplication by the hessian,''
\newblock {\em Neural Computation}, vol. 6, no. 1, pp. 147--160, 1994.

\bibitem{avron2011randomized}
Haim Avron and Sivan Toledo,
\newblock ``Randomized algorithms for estimating the trace of an implicit symmetric positive semidefinite matrix,''
\newblock {\em J. ACM}, vol. 58, no. 2, pp. 8:1--8:34, 2011.

\bibitem{amari1998natural}
Shun ichi Amari,
\newblock ``Natural gradient works efficiently in learning,''
\newblock {\em Neural Computation}, vol. 10, no. 2, pp. 251--276, 1998.

\bibitem{halko2011finding}
Nathan Halko, Per-Gunnar Martinsson, and Joel~A. Tropp,
\newblock ``Finding structure with randomness: Probabilistic algorithms for constructing approximate matrix decompositions,''
\newblock {\em SIAM Review}, vol. 53, no. 2, pp. 217--288, 2011.

\bibitem{hendrycks2021mmlu}
Dan Hendrycks, Collin Burns, Steven Basart, Andy Zou, Mantas Mazeika, Dawn Song, and Jacob Steinhardt,
\newblock ``Measuring massive multitask language understanding,''
\newblock {\em arXiv preprint arXiv:2009.03300}, 2021.

\bibitem{cobbe2021gsm8k}
Karl Cobbe, Vineet Kosaraju, Mohammad Bavarian, Jacob Hilton, Reiichiro Nakano, Christopher Hesse, and John Schulman,
\newblock ``Training verifiers to solve math word problems,''
\newblock in {\em NeurIPS}, 2021.

\bibitem{clark2018arc}
Peter Clark, Isaac Cowhey, Oren Etzioni, Tushar Khot, Ashish Sabharwal, Carissa Schoenick, and Oyvind Tafjord,
\newblock ``Think you have solved question answering? try arc, the ai2 reasoning challenge,''
\newblock in {\em NAACL-HLT}, 2018, pp. 47--56.

\bibitem{zellers2019hellaswag}
Rowan Zellers, Ari Holtzman, Yonatan Bisk, Ali Farhadi, and Yejin Choi,
\newblock ``Hellaswag: Can a machine really finish your sentence?,''
\newblock in {\em ACL}, 2019, pp. 4791--4800.

\bibitem{clark2019boolq}
Christopher Clark, Kenton Lee, Ming-Wei Chang, Tom Kwiatkowski, Michael Collins, and Kristina Toutanova,
\newblock ``Boolq: Exploring the surprising difficulty of natural yes/no questions,''
\newblock in {\em NAACL-HLT}, 2019, pp. 2924--2936.

\bibitem{sakaguchi2021WinoGrande}
Keisuke Sakaguchi, Ronan~Le Bras, Chandra Bhagavatula, and Yejin Choi,
\newblock ``Winogrande: An adversarial winograd schema challenge at scale,''
\newblock in {\em AAAI}, 2021, pp. 8732--8740.

\bibitem{gao2021lmeval}
Leo Gao, Stella Biderman, Sid Black, Laurence Golding, Travis Hoppe, Charles Foster, Jason Phang, Horace He, Anish Thite, Noa Nabeshima, Shawn Presser, and Connor Leahy,
\newblock ``A framework for few-shot language model evaluation,''
\newblock {\em arXiv preprint arXiv:2108.12470}, 2021.

\end{thebibliography}
}
\end{document}